\documentclass[11pt,a4paper]{article}
\usepackage[hyperref]{acl2017}
\usepackage{times}
\usepackage{latexsym}

\usepackage{url}

\usepackage{latexsym}
\usepackage[]{xcolor}

\usepackage{array}
\usepackage{algorithm}
\usepackage[noend]{algpseudocode}
\usepackage{algorithmicx}
\usepackage{amssymb}
\usepackage{amsmath}
\usepackage{arydshln}
\usepackage{multirow}
\usepackage{tabularx}
\usepackage{caption}
\usepackage{multirow}

\title{Automatic Identification of AltLexes using Monolingual Parallel Corpora} 

\aclfinalcopy

\author{Elnaz Davoodi \and Leila Kosseim\\
    Department of Computer Science and Software Engineering\\
Concordia University\\
1515 Ste-Catherine Blvd. West \\
Montreal, Quebec, Canada, H3G 2W1\\
  {\tt ed.davoodi@gmail.com} \\
  {\tt leila.kosseim@concordia.ca}}

\date{}

\begin{document}
\maketitle

\begin{abstract}

The automatic identification of discourse relations is still a challenging task in natural language processing. Discourse connectives, such as \textit{since} or \textit{but}, are the most informative cues to identify explicit relations; however discourse parsers typically use a closed inventory of such connectives. As a result, discourse relations signaled by markers outside these inventories (i.e. AltLexes) are not detected as effectively. In this paper, we propose a novel method to leverage parallel corpora in text simplification and lexical resources to automatically identify alternative lexicalizations that signal discourse relation. When applied to the Simple Wikipedia and Newsela corpora along with WordNet and the PPDB, the method allowed the automatic discovery of 91 AltLexes.
\end{abstract}

\section{Introduction}

Understanding a text goes beyond understanding its textual units in isolation; the relation between these units must also be understood.  Discourse connectives such as \textit{since}, \textit{but}, etc. are often used to explicitly connect textual units and signal the presence of specific \textit{explicit} discourse relations such as \textsc{contrast}, \textsc{cause}, etc. The Penn Discourse Tree Bank (PDTB) framework \cite{Prasad08} defined a closed list of discourse connectives. \textit{Implicit}, \textit{AltLex} and \textit{EntRel} are other realizations of discourse relations in the PDTB. AltLex (or alternative lexicalization relations), which are understudied in computational discourse processing, are signalled using an open list of lexical markers that are not part of the PDTB inventory of discourse connectives.


\begin{figure*}[]
    \centering
\small
\begin{tabular}{p{7.5cm}|p{7.5cm}}
    \hline
      \multicolumn{1}{c|}{\textbf{Complex}}   & \multicolumn{1}{c}{\textbf{Simple} }\\ \hline \hline
      These works he produced and published himself, \textbf{whilst} his much larger woodcuts were mostly commissioned work. [Non-Explicit] & He created and published his works himself, \textbf{but} his larger works were mostly commissioned work to be sold. [Explicit \textsc{contrast}] \\ \hline
    \end{tabular}

    \caption{\small{No explicit relation is detected in the complex sentence (left), but an explicit \textsc{contrast} relation is identified in the simple sentence (right). The example is taken from the Simple English Wikipedia corpus~\cite{coster2011learning}}}

    \label{example_1}
\end{figure*}

Figure \ref{example_1} shows a pair of sentences that convey the same information; however only one sentence contains a discourse connective from the PDTB inventory. Hence, a discourse parser using the PDTB inventory of connectives would easily identify the explicit \textsc{contrast} relation in the first sentence but will likely not tag the second sentence because \textit{whilst} is not part of the PDTB inventory of discourse connectives.  
 However, the writer's intention can be understood using other means such as the use of an alternative lexical marker (i.e. AltLex), a change of tense, a structural signal, etc. Thus, discourse parsers can benefit from the automatic identification of AltLexes that can signal discourse relations. 

 According to ~\citet{pitler2009}, discourse connectives constitute strong clues to detect explicit relations, hence discourse parsers have relied on them as valuable features in order to identify explicit discourse relations automatically~\cite{LinNK14}. Similarly, the presence of alternative lexical markers is a strong indicator of an AltLex relation; however since the list of such markers is open, identifying them is a challenge.

 \section{Background}

Discourse connectives \cite{blakemore1987,knott1994,schourup1985} are the most informative signals of explicit discourse relations~\cite{pitler2009}. However, they are not well-defined in linguistics. \citet{levinson1983} defined discourse connectives as words and phrases such as \textit{after all}, \textit{actually}, etc. that connect an utterance to the prior discourse. \citet{zwicky1985} considered discourse connectives as a class of particles, but did not specify what particles are considered as discourse connectives. \citet{schiffrin1988} also defined discourse connectives as words that connect dependent textual units in a discourse. According to Schiffrin, discourse connectives do not belong to any linguistic class and except for a few discourse connectives such as \textit{oh} and \textit{well}, most carry meaning. \citet{redeker1991} revised this definition; even though she agreed that discourse connectives have meaning by themselves, she argued that they should contribute to the semantic interpretations of the discourse. Apart from research efforts aiming at defining discourse connectives, another line of research has focused on providing a list of discourse connectives in English \cite{Prasad08,andersen2001,blakemore2002,fischer2000,sanders1992,knott1996} and other languages \cite{pasch2003,travis2005}. While most of these inventories have been built by hand, some work has attempted to identify them automatically. \citet{laali2014inducing} used the Europal parallel corpus and collocation techniques to induce French connectives from their English counterparts. Following this work, \citet{hideyidentifying} built a parallel corpus of causal and non-causal AltLexes using word alignment with PDTB discourse connectives as initial seeds. Our work is different from these as we use already existing parallel corpora in text simplification and extract discourse information automatically using the state of the art discourse parser. In addition, instead of focusing on only one relation, we generalize the problem to all PDTB relations. We also use external resources which are shown to have advantages over word alignment \cite{versley2010discovery} in similar tasks. Lastly, the PDTB AltLexes only captures inter-sentence relations. Our contribution overcomes this limitation by identifying intra-sentence relations.

\section{Identification of Discourse Connectives}


In order to automatically identify AltLexes, we used the notion of text simplification~\cite{siddharthan2014survey,kauchak2013}. While two texts may convey the same meaning, they may exhibit different complexity levels~\cite{pitler2008,davoodi-kosseim:2016:SIGDIAL}. This difference in complexity level may be the result of various linguistic choices; at the lexical level (e.g. using frequent vs. abandoned words), the syntactic level (e.g. using active vs. passive voice) or even the discourse level (e.g. using an implicit vs. an explicit discourse relation). The main assumption in text simplification is that it is possible to reduce a text's complexity while preserving its meaning as much as possible. Because discourse relations are semantic in nature, we can therefore assume that they are also preserved as much as possible during text simplification. However, the lexical realization of discourse relations (i.e. explicit versus non-explicit) or the choice of a discourse connective (e.g. \textit{but} vs. \textit{however}) may change. The removal of a discourse relation may happen if the discourse argument is considered non-essential. For example, Figure \ref{removal} shows a pair of aligned sentences where the complex version contains an explicit \textsc{synchrony} relation signalled by \textit{when}; while the discourse argument and consequently the explicit discourse connective has been removed in the simple version. Hence, given a sentence and its simplified version, three phenomena can occur:

\begin{tabular}{lp{7.1cm}}
\hspace*{-.6cm}1. & \hspace*{-.43cm}a discourse connective is replaced by another (e.g. \textit{although} $\Rightarrow$ \textit{though}), \\
\hspace*{-.6cm}2.&  \hspace*{-.43cm}a discourse connective is replaced by another lexical choice (i.e. word or phrase) which is not considered as a discourse connective in the inventory used (e.g. \textit{although} $\Rightarrow$ \textit{despite}), or \\
\hspace*{-.6cm}3.&  \hspace*{-.43cm}a discourse connective is removed completely. 
\end{tabular}

In cases (1) and (2) above, the discourse relation is preserved, while in case (3) the discourse relation is either removed or changed to an implicit relation. The focus of this paper is to study case (2) and use such a phenomenon to automatically identify AltLexes. 

\begin{figure}[]
    \centering
    \small
    \begin{tabular}{|l|}
    \hline
      \multicolumn{1}{|p{7.5cm}|}{ \textbf{Complex: }\textbf{When} the show was broadcast, Rupert Boneham won the million dollars. [Explicit \textsc{synchrony}]}\\ 
        
 \multicolumn{1}{|p{7.5cm}|} {\textbf{Simple: }Rupert Boneham won the million dollars.} \\ \hline

    \end{tabular}

  \caption{\small{An example of the removal of a discourse argument and consequently the removal of a discourse relation.}}

  \label{removal}
\end{figure}

\subsection{Data Sets}

To discover AltLexes automatically, we created two sentence-aligned data sets using standard corpora in text simplification. The first data set was created from the Simple English Wikipedia corpus~\cite{coster2011learning}; the other was created from the Newsela corpus~\cite{newsela2015problems}.


The Simple English Wikipedia (SEW) corpus~\cite{coster2011learning} contains two sections: 1) article-aligned and 2) sentence-aligned. Here, we used the sentence-aligned section, which contains 167,686 pairs of aligned sentences.




In order not to overfit to a specific corpus, we also used the Newsela (News) corpus~\cite{newsela2015problems}. This corpus contains 1,911 English news articles which have been manually re-written at most 5 times, each time with decreasing complexity level. We used this article-aligned corpus to align it at the sentence-level using an approach similar to \cite{coster2011learning}. Then, two native English speakers evaluated the alignments. The Kappa inter-annotation agreement was 0.898 computed on 100 randomly chosen alignments.  

\subsection{Methodology}

According to the PDTB framework, each Altlex can be substituted with at least one discourse connective \citep{Prasad08}. Based on this, to discover AltLexes automatically, we first parsed both sides of the aligned sentences of both data sets to extract discourse information. This was done using the PDTB-style End-to-End parser~\cite{LinNK14}. This parser was selected as it is currently the best performing parser to identify explicit relations, with an F-measure between 80.61\% and 86.77\% depending on the evaluation criteria. Because it uses the PDTB framework, the parser  uses the inventory of 100 discourse connectives from the PDTB. The result of this tagging was categorized into one of the following cases:



\noindent\begin{tabular}{lp{2.65in}}
1.  &  \hspace*{-.4cm} \textit{NonExp-NonExp}: a non-explicit\footnotemark~discourse relation occurs in both sentences. \\
 \end{tabular}
 \begin{tabular}{lp{2.65in}}
2. &  \hspace*{-.4cm}  \textit{Exp-Exp}: the same discourse relation and discourse connective occur in both sentences.\\
 \end{tabular}
 \begin{tabular}{lp{2.65in}}
 3.& \hspace*{-.4cm} \textit{NonExp-Exp}: a non-explicit relation occurs in the complex sentence, but an explicit one is used in the simple sentence.\\

 4.& \hspace*{-.4cm} \textit{Exp-NonExp}: an explicit relation occurs in the complex sentence, but no relation is used in the simple sentence.  \\
 \end{tabular}
 \begin{tabular}{lp{2.65in}}
 
 5.  & \hspace*{-.4cm} \textit{Other}: \\
  & \hspace*{-.4cm}a. \textit{Same Relation-Different Connective}: the same explicit relation is used but with different discourse connectives in both sentences.\\
 & \hspace*{-.4cm}b. \textit{Different Relation-Different Connective}: a different explicit relation and a different discourse connective are used. \\
 & \hspace*{-.4cm}c. other cases including several explicit relations within a single sentence.\\
 \end{tabular}

\begin{table}[]
    \centering
\scriptsize
    \begin{tabular}{|l||r|r||r|r|} 
    \hline  

    \textbf{Change} & \multicolumn{2}{c||}{\textbf{SEW-based DS}}& \multicolumn{2}{c|}{\textbf{News-based DS}} \\ \hline\hline    
     (1) NonExp-NonExp & 116,852 & 69.68\% & 18,384 & 50.00\% \\
     (2) Exp-Exp & 19,735 & 11.76\% & 2,660 & 7.23\%\\
     \textbf{(3) NonExp-Exp} & \bf 7,868 &\bf 4.69\% & \bf 1,129 & \bf 3.07\%\\
     \textbf{(4) Exp-NonExp} & \bf 9,490 &\bf 5.65\% & \bf 1,733 &\bf 4.71\%\\
     (5) other  & 13,741 & 8.22\% & 12,862 & 34.99\%\\\cdashline{1-5}
     Total & 167,686 & 100\% & 36,768 & 100\% \\ \hline
     \end{tabular}

    \caption{\small{Frequency of the discourse changes across complexity levels in the SEW-based and News-based data sets.}}

    \label{change}
\end{table}

\begin{table*}
    \centering
    \scriptsize
      \begin{tabular}{|l||r|r|r||r|r|r||r|r|}
        \hline
    & \multicolumn{3}{c||}{\textbf{SEW-based Data set}} 
    & \multicolumn{3}{c||}{\textbf{News-based Data set}} 
    & \multicolumn{2}{c|}{\textbf Overall}\\ \cline{2-9}
   
       \textbf{Discourse} 
       & &\multicolumn{2}{c||}{\textbf New AltLexes from} 
       &  &\multicolumn{2}{c||}{\textbf New AltLexes from}    &   &{\textbf New markers from}  \\ 
       
    \textbf{Relation} &  \bf Alignments &\bf PPDB & \bf WordNet &   \bf Alignments  & \bf PPDB  & \bf WordNet &  \bf Alignments  &\textbf{SEW $\cup$ News} \\ \hline \hline
    
      \textsc{Asynchronous} & 2,561 & 15  & 3 & 327 & 6 & 2 & 2,888  & 20 \\
    \textsc{Synchrony} & 1,990 & 2  & 1 & 395 & 0 & 0 & 2,385 & 3 \\
    \textsc{Cause} & 1,359& 18  & 1 & 256& 3 & 0 & 1,615 & 19 \\
    \textsc{condition} &296 & 0  & 0 & 141 & 1 & 0 & 437  & 1\\
    \textsc{contrast} & 2,568 & 6  & 1 & 667 & 5 &  6 & 3,235 &  9 \\
    \textsc{concession} & 393 & 3 & 0 & 64 & 0 & 0 &  457 & 3\\
    \textsc{conjunction} & 7,738 & 25  & 1 &914 & 12 & 3 & 8,652  & 27  \\
    \textsc{instantiation} & 159 & 3  & 1 & 33 & 0 & 0 & 192  & 3 \\
    \textsc{restatement} &63 & 1 & 0  & 13 & 0 & 0 & 76 & 1 \\
    \textsc{alternative} & 220 & 5 & 0 & 51 & 1 & 0 & 271 & 5  \\
    \textsc{exception} & 8 & 0  & 0 & 0 & 0 & 0 & 8 & 0 \\
    \textsc{list} & 3 & 0 & 0 & 1 & 0 & 0 & 4 & 0 \\ \cdashline{1-9}
    \textbf{Total} & \bf 17,358 & \bf 79 & \bf 8 & \bf 2,862 & \bf 28 & \bf 11 & \bf 20,220  &\bf 91 \\
    
    \hline
  
  \end{tabular}

    \caption{\small{Number of {\em Exp-NonExp} and {\em NonExp-Exp} alignments and newly identified AltLex types.}}

    \label{stat_2}
\end{table*}

\footnotetext{A non-explicit discourse relation can refer to an implicit, and AltLex discourse relation or to no discourse relation.}

\begin{figure*}[!htbp]
    \centering
\small
\begin{tabular}{p{7.6cm}|p{7.7cm}}
    \hline
      \multicolumn{1}{c|}{\textbf{Complex}}   & \multicolumn{1}{c}{\textbf{Simple} }\\ \hline \hline
      Now they have drones in 15 states, including California and Texas. \textbf{Before} they started the business, the two covered fields on foot or in vehicles. [Explicit \textsc{asynchronous}]& Now they have drones in 15 states, including California and Texas. Fiene \underline{used to} check farm fields on foot or with vehicles.  \\ \hline
    \end{tabular}

    \caption{\small{An example of discourse marker and potential AltLex having different syntactic class across complexity levels.}}

    \label{change_syntax}
\end{figure*}

Table~\ref{change} shows the frequency of these transformations in the two data sets. To discover AltLexes, we only considered cases (3) and (4), where only one side of the alignment includes a PDTB discourse connective. This gave rise to a total of 20,220 aligned sentences. We then used two external resources: 1) the paraphrase database (PPDB)~\cite{ganitkevitch2013ppdb} and 2) WordNet~\cite{miller1995wordnet}. The PPDB comes in six sizes from $S$ to $XXXL$. The smaller versions of the PPDB contain more precise paraphrases with higher confidence scores; while the larger versions have more coverage. We choose the PPDB version L to have a good compromise between the precision and coverage of paraphrases. We took the discourse connective from the explicit side and looked for an alternative lexicalization (a synonym or paraphrase) in the external resources. If any of its alternative lexicalization appeared in the non-explicit side, we considered it as an AltLex to signal the relation. We then replaced the AltLex with the discourse connective from the explicit side and parsed the new sentence with the PDTB-style End-to-End parser again. This process is shown in Figure~\ref{example_case2}. On average, each discourse connective was replaced by 23.2 alternative lexicalizations taken from the PPDB and 12.3 from WordNet.


If the parser detected the same relation (see Figure~\ref{example_case2}), then the potential marker was considered as an AltLex. On the other hand, because the End-to-End discourse parser uses both the discourse connective and syntactic features, if it was not capable of detecting the relation in the replaced sentences, we concluded that either (1) the relation existed, but the parser could not detect it, (2) the AltLex does not signal the discourse relation or (3) the relation does not exist (see Figure \ref{example_case3}). Because we did not use any syntactic filter, the replacement of the discourse connective may alter the syntax of the sentence such that the parser is unable to detect the relation. This is why, regardless of the reason, if the parser was not able detect the relation, we discarded the AltLex.

\section{Results and Discussion}

\label{analysis}

Table \ref{stat_2} shows the number of sentence alignments mined and the number of potential AltLexes (i.e. type count) identified in each data set for each level 2 PDTB relation. Overall, by mining 17,358 NonExp-Exp and Exp-NonExp alignments, the SEW-based data set allowed the discovery of 79 AltLexes from the PPDB and 8 from WordNet; whereas, the News-based data set, providing only 2,862 alignments, allowed the discovery of 28 AltLexes from PPDB and 11 from WordNet. Using both corpora and both lexical resources, the method found 91 AltLexe tokens, which account for 533 AltLexes. It is interesting to note that, overall, the approach did not find any alternate lexicalizations for some relations such as \textsc{list} or \textsc{exception} and only one for \textsc{condition}. It is not clear if this is because these relations are typically signalled using a rather fixed inventory of discourse markers or because of the low number of such alignments. Indeed, in the PDTB, out of 624 tagged AltLex relations, only 6 are labeled as \textsc{restatement}, 1 as \textsc{exception} and 2 as \textsc{condition}. 

\begin{figure}
\centering

    \centering
    \small
    \begin{tabular}{|l|}
    \hline
      \multicolumn{1}{|p{7.2cm}|}{ \textbf{Complex: }It's a very special place \textbf{because} this site, this area, has been tied to the history and life of African-Americans since about the early 1800s. [Explicit \textsc{cause}] }\\ 
        
 \multicolumn{1}{|p{7.2cm}|}{\textbf{Simple: }It has been tied to the history and life of African-Americans \textit{since} [\textsc{synonym of because}] about the early 1800s.} 
  \\
    \multicolumn{1}{|p{7.2cm}|}{\centering$\Downarrow$}
  
  \\
 \multicolumn{1}{|p{7.2cm}|}{ \textbf{Simple after substitution: }It has been tied to the history and life of African-Americans \textbf{because} about the early 1800s.} \\ \hline

    \end{tabular}

  \caption{\small{An example which does not lead to an AltLex. After the substitution, the \textsc{cause} relation is still not identified.}}

  \label{example_case3}
\end{figure}

 \begin{figure}     
    \centering
    \small
    \begin{tabular}{|l|}
    \hline
      \multicolumn{1}{|p{7.5cm}|}{\textbf{Complex: }Today, the comic arm of the company flourishes \textit{despite} [\textsc{synonym of though}] no longer having its own universe of super powered characters.}\\

 \multicolumn{1}{|p{7.5cm}|}{\textbf{Simple: }Today, the company does very well even \textbf{though} they do not have their own universe of super powered characters. [Explicit \textsc{contrast}] } \\

    \multicolumn{1}{|p{7.5cm}|}{\centering$\Downarrow$}\\

 \multicolumn{1}{|p{7.5cm}|}{ \textbf{Complex after substitution: }Today, the comic arm of the company flourishes \textbf{though} no longer having its own universe of super powered characters. [Explicit \textsc{contrast}]} \\ \hline

    \end{tabular}

  \caption{\small{An example which leads to an AltLex. After the substitution, the \textsc{contrast} relation is identified.}}

  \label{example_case2}

\end{figure}

On the other hand, relations such as \textsc{conjunction}, \textsc{asynchronous} and \textsc{cause} provided a large number of alignments from which we identified a variety of AltLexes. For example, the PPDB identified ``\textit{caused by}'', ``\textit{resulting}'', ``\textit{causing}'', ``\textit{this being so}'', etc. as AltLexes to signal a \textsc{cause} relation. The full inventory of the automatically identified AltLexes can be found at: \url{http://Anynomous}.

In addition, as can be seen in Table \ref{stat_2}, the number of potential AltLexes coming from the PPDB is higher than the number of AltLexes coming from WordNet. This may be due to the difference of the coverage of these two resources as WordNet is smaller than the PPDB. Another possible reason is that each word in PPDB has a list of paraphrases with various syntactical classes. Thus, if the syntactic class of a discourse marker is changed in the simplification process, it is more probable that the PPDB covers more syntactical variations of the discourse marker compared to WordNet. Figure \ref{change_syntax} shows an example taken from the Newsela corpus. In this example, the discourse marker \textit{before} in the complex version signals \textsc{asynchronous} relation, but is tagged as subordinating conjunction. In the paraphrase database, \textit{used to} is one of the paraphrases of discourse marker \textit{before}. In the simple version of this example, the verb \textit{used to} is signalling the same relation (i.e. \textsc{asynchronous} relation) which is captured as an AltLex.

\section{Future Work}

As future research, we plan to assign a confidence level to the automatically identified AltLexes by using a syntactic filter to replace potential markers only if they lead to syntactically correct sentences and by using their frequency of appearance in the parallel corpora. Another interesting line of research would be to evaluate if discourse parsers can increase their performance using such new list. 

\section*{Acknowledgement}
The authors would like to thank the anonymous reviewers for their feedback on an earlier version of the paper. This work was financially supported by the Natural Sciences and Engineering Research Council of Canada (NSERC).

\bibliographystyle{acl}

\end{document}